\documentclass[11pt]{article}

\usepackage[preprint]{acl}

\usepackage{times}
\usepackage{latexsym}

\usepackage[T1]{fontenc}

\usepackage[utf8]{inputenc}

\usepackage{microtype}

\usepackage{inconsolata}

\usepackage{graphicx}
\usepackage{listings}
\usepackage{booktabs}
\usepackage{multirow}

\title{A Systematic Approach for Large Language Models Debugging}

\author{
  Basel Shbita\textsuperscript{1},
  Anna Lisa Gentile\textsuperscript{1},
  Bing Zhang\textsuperscript{1},
  Sungeun An\textsuperscript{1},
  Shailja Thakur\textsuperscript{2},\\
  \textbf{ 
  Shubhi Asthana\textsuperscript{1},
  Yi Zhou\textsuperscript{1},
  Saptha Surendran\textsuperscript{2},
  Farhan Ahmed\textsuperscript{1},
  Rohan Kulkarni\textsuperscript{2},}\\
  \textbf{ 
  Yuya Jeremy Ong\textsuperscript{1},
  Chad DeLuca\textsuperscript{1},
  Hima Patel\textsuperscript{2}}\\
  \textsuperscript{1}IBM Research, San Jose, CA, USA \\
  \textsuperscript{2}IBM Research, Bengaluru, India \\
  \small\texttt{\{basel, annalisa.gentile, bing.zhang, sungeun.an, shailja.thakur\}@ibm.com},
  \texttt{sasthan@us.ibm.com},\\
  \small\texttt{\{yi.zhou, saptha.surendran, farhan.ahmed, rohan-kulkarni7, yuyajong\}@ibm.com},\\
  \small\texttt{delucac@us.ibm.com, himapatel@in.ibm.com}
}

\begin{document}
\maketitle

\begin{abstract}

Large language models (LLMs) have become central to modern AI workflows, powering applications from open-ended text generation to complex agent-based reasoning.
However, debugging these models remains a persistent challenge due to their opaque and probabilistic nature and the difficulty of diagnosing errors across diverse tasks and settings.
This paper introduces a systematic approach for LLM debugging that treats models as observable systems, providing structured, model-agnostic methods from issue detection to model refinement.
By unifying evaluation, interpretability, and error-analysis practices, our approach enables practitioners to iteratively diagnose model weaknesses, refine prompts and model parameters, and adapt data for fine-tuning or assessment, while remaining effective in contexts where standardized benchmarks and evaluation criteria are lacking.
We argue that such a structured methodology not only accelerates troubleshooting but also fosters reproducibility, transparency, and scalability in the deployment of LLM-based systems.
\end{abstract}
\section{Introduction}

Large language models (LLMs) have rapidly advanced the capabilities of artificial intelligence, enabling breakthroughs in natural language understanding and reasoning, and achieving impressive performance in a wide range of tasks~\citep{naveed2025comprehensive, matarazzo2025survey, yang2024harnessing, brown2020language, wang2019superglue}.
LLMs underpin a wide range of workflows, from domain-specific fine-tuning~\citep{ouyang2022training} pipelines to agent-based systems capable of multi-step planning and tool use~\citep{wu2025agentic}.
Despite their success, LLMs present unique challenges: their outputs are inherently probabilistic, their behavior often opaque, and their ``failures'' difficult to isolate~\citep{gao2025llm}.
As reliance on LLMs grows, the ability to systematically identify, analyze, and resolve issues becomes a critical requirement.

Traditional debugging practices fall short in addressing the complexities of LLMs.
Metrics designed for well-defined tasks rarely capture the nuances of open-ended generation, where multiple valid outputs exist~\citep{chang2024survey, li2024llms, wang2025can}.
Benchmark-driven evaluation, while useful, is often limited in scope and may not reflect the conditions under which models are deployed.
In emerging or low-resource domains, standardized benchmarks may be absent altogether, making \textit{ad-hoc} debugging insufficient and obscuring the root causes of model failures.

This paper proposes a systematic approach for LLM debugging that treats models as observable systems, enabling structured and model-agnostic workflows for issue detection, interpretability, and error analysis.
By unifying these practices into an integrated framework, practitioners can iteratively diagnose weaknesses, refine prompts, and adapt data for fine-tuning or evaluation, even in environments where conventional resources such as evaluation data are unavailable.
This approach standardizes debugging as a reproducible, scalable engineering practice applicable across tasks, domains, and deployment settings.

Our key contributions are threefold:
(i) we introduce a four-phase debugging framework for diagnosing and improving LLM performance, structured around the stages of issue detection, evidence gathering, behavioral analysis, and iterative refinement,
(ii) we demonstrate the applicability of this framework across three major categories of tasks: well-defined tasks, underspecified tasks, and agentic or workflow-level applications, and
(iii) we ground the framework in a set of representative use cases drawn from practical LLM development scenarios, showing how it enables systematic debugging and refinement in diverse settings.

\section{Related Work}

Recent work on LLM evaluation emphasizes moving beyond traditional metrics such as accuracy, BLEU, and ROUGE~\citep{chang2024survey}.
For open-ended generation, exact-match and string-based metrics are brittle, and existing evaluation paradigms often fail to cover real-world conditions or adapt to new domains~\citep{srivastava2023beyond, ni2025survey}.
To address these challenges, LLM-as-a-Judge (LLMaJ) approaches inject semantic reasoning into evaluation~\citep{zheng2023judging, li2024llms}, but introduce concerns around bias and consistency~\citep{li2025evaluating}.
Beyond single-turn interactions, AI agents present distinct debugging challenges, as failures cascade across trajectories with complex prompts, multi-step reasoning, and dynamic tool use~\citep{majdoub2025adaptivesoftwareagentsdebugging}.
Existing techniques include self-verification, reflection-based methods~\citep{huang2022large, shinn2023reflexion}, and supervised approaches~\citep{xiang2025self}.
However, no existing framework forms a closed loop from behavioral analysis to prompt redesign and data augmentation in agentic settings.
Our work bridges this gap by introducing a unified, model-agnostic framework that operationalizes LLM debugging as an engineering practice.

\section{Systematic Debugging of LLMs}
\label{sec:approach}



\paragraph{Issue Detection.}
As shown in Figure~\ref{fig:flow_diagram}, debugging begins when a model produces outputs that are unsatisfactory or deviate from expectations, which can manifest as drops in benchmark performance, inconsistent behavior across similar prompts, or failures in reasoning and workflow execution.
The goal at this stage is to characterize the issue in practical terms so it can be systematically addressed, rather than relying on \textit{ad-hoc} observations.
If benchmarks exist, they provide the quickest entry point: profiling across standard suites in the Open LLM Leaderboard~\citep{leaderboard-v1, leaderboard-v2} or specialized benchmarks such as RULER~\citep{hsieh2024rulerwhatsrealcontext} for context length or ALERT~\citep{tedeschi2024alert} for safety often reveals systematic weaknesses.
Crucially, issue detection is not about fully diagnosing root causes but about localizing where failures are most visible, creating a focused starting point for evidence gathering and analysis.

\begin{figure*}[t]
  \centering
  \includegraphics[width=1\linewidth]{imgs/llm-debug-flow-draft-04.png}
  \caption{Overview of our systematic debugging workflow. After detecting an issue, the process branches based on the available evidence, including benchmarks, task definition, and evaluation criteria. Structured analyses then guide refinements to prompts, hyperparameters, evaluation protocols, or training data, with each cycle feeding back into re-testing and continued iteration.}
  \label{fig:flow_diagram}
\end{figure*}


\paragraph{Evidence Gathering.}
After an issue has been detected, the next step is to ensure that sufficient evidence is available to support structured analysis.
This phase is about assembling the necessary prerequisites, including evaluation data, criteria, and metrics, which will make later behavioral analysis meaningful.
The first question is whether evaluation data or benchmark sets are available.
If such resources exist, they form the natural starting point.
When they are absent, practitioners should construct a small evaluation set manually and, if necessary, expand it synthetically to ensure a minimal evidence base.
One effective strategy here is to employ \textit{Scalable Synthetic Data Generation (SDG)}~\cite{fms-dgt}, an open-source general-purpose LLM-wrapper framework designed for producing large quantities of synthetic data using in-context examples.

For agentic workflows or tool-calling, trajectories become the key evidence: agent runs, tool calls, and reasoning steps should be logged and broken into interpretable units such as individual tool calls or reasoning segments, enabling isolation of recurring failure modes and alignment with frameworks such as ReAct~\citep{yao2023react}.
For non-agentic cases, the relevant question is whether the task is well-defined.
For tasks with unambiguous correctness criteria (e.g., multiple-choice QA or exact string match), existing evaluation data can be applied directly.
For tasks that are underspecified, however, evaluation criteria or metrics may be missing.
In such cases, practitioners should construct lightweight evaluators.
Code-based validators can check for structural correctness, while LLMaJ rubric-driven criteria can approximate semantic adequacy.
Even when approximate, evaluators ensure that the evidence gathered is systematic rather than anecdotal.
At this point, practitioners decide whether the evidence base is sufficient or whether additional data or criteria must be gathered before proceeding with structured analysis.


\paragraph{Behavioral Analysis.}
In this phase, the focus shifts to behavioral analysis, where collected evidence is examined to better understand model behavior and its sources of error.
The process follows a small number of decision points that guide which analyses are most informative.
When evaluation results are available, they can be profiled systematically along semantic (e.g., categories, domains) and structural (e.g., input length, format) dimensions; otherwise, practitioners collect new runs, outputs, prompts, or trajectory metadata to build an evidence base for analysis.
The next step is to determine which type of analysis is most suitable.
In settings where only external behavior is observable, a \textit{glass-ceiling test} can identify data gaps: adding evaluation data to the training set and re-evaluating reveals whether failures stem from insufficient data coverage or deeper architectural limitations.

When available, practitioners can assess token-level diagnostic signals, such as logits or statistical measures including entropy, varentropy, confidence, and surprisal, which quantify model uncertainty and can reveal anomalies such as hesitation or overfitting to narrow patterns.
By indicating whether confidence is well-calibrated or systematically skewed, these signals provide a quantitative view of sequence processing and can guide refinement by highlighting where targeted data augmentation or diagnostic fine-tuning is most likely to yield improvements.
Beyond static analysis, fine-tuning itself can be used diagnostically: by adding targeted data and observing behavioral changes under different regimes—such as Low-Rank Adaptation (LoRA)~\cite{hu2022lora}, quantized LoRA (QLoRA)~\citep{dettmers2023qlora}, activated LoRA (ALoRA)~\citep{greenewald2025activated}, or full fine-tuning with varied data mixtures—practitioners can probe whether failures stem from data coverage gaps or deeper architectural limitations, enabling more principled refinements in the next phase.


\paragraph{Iterative Refinement.}
Insights from behavioral analysis guide corrective strategies in the final phase of the debugging cycle.
The objective is both to mitigate observed failures and to test hypotheses about their causes.
Interventions typically target three areas: content, stability, and data.
Content issues can be addressed through prompt refinement, adjusting structure or examples to steer outputs.
Stability issues often respond to hyperparameter tuning, such as modifying decoding strategies or temperature.
When failures reflect gaps in evaluation or training data, updating evaluation protocols or adapting data through curation, augmentation, or fine-tuning provides the most direct path to improvement.
This phase is not a terminal step but a loop: refinements are introduced, the model is re-tested, and new evidence is gathered.
The cycle continues until performance reaches an acceptable level of accuracy, reliability, and stability.



\section{Use Cases}
\label{sec:use_cases}


\subsection{Well-Defined Tasks}
\label{sec:welldefined_use_cases}

We refer to \textit{well-defined tasks} as those where
(i) evaluation criteria are precise,
(ii) correctness can be judged unambiguously, and
(iii) evaluation benchmarks are available.
Debugging tends to focus on systematic weaknesses in the data or model configuration, with fine-tuning often providing effective remedies.


\paragraph{Natural Language to GraphQL (NL2GQL).}
We evaluated the task of generating GraphQL~\citep{quina2023graphql} queries, a well-defined problem with clear correctness criteria:
(i) generate the query from a natural language description,
(ii) execute against a database, and
(iii) verify the result.
Unlike text-to-SQL~\citep{zhang2023sciencebenchmarkcomplexrealworldbenchmark, zhang2024benchmarkingtexttosqlcapabilitylarge, gao2023texttosqlempoweredlargelanguage, ma2024evaluatingllmstexttosqlgeneration}, GraphQL is less studied but has a public benchmark~\citep{kesarwani-etal-2024-graphql} that provides 10K labeled training samples.
Using \textit{Granite-3.1-8B-Instruct}~\citep{granite2024granite} as a test model, we found initial accuracy around 20\%, with prompt optimization raising it only to 27\%.
A representative failure case is provided in Appendix~\ref{sec:appndx_nl2gql}.
A \textit{glass-ceiling test} confirmed the gap was due to lack of training exposure, since the model reached 96\% accuracy once evaluation data were included in training.
We then proceeded to fine-tune the models in two different ways:
(i) \textit{vanilla fine-tuning}, with a small (10K) amount of manually curated labeled data, obtained from~\cite{kesarwani-etal-2024-graphql}: this improved accuracy by \textasciitilde30\%, achieving an overall 50\% accuracy, and
(ii) data-mixture fine-tuning approach, where the same 10K GraphQL samples were injected into a larger (\textasciitilde1M) generic and technical (math, coding, etc.) labeled data: this  significantly improved performance by \textasciitilde40\%, achieving an overall 60\% accuracy for the task.
We also verified that the fine-tuning has no negative effect on the standard base models' capabilities, i.e. does not generate catastrophic forgetting on other generic tasks.


\paragraph{Temporal Reasoning.}
Another pervasive challenge is observed with respect to complex temporal tasks~\citep{wei2025timemultilevelbenchmarktemporal, wang2024trambenchmarkingtemporalreasoning, tan2023benchmarkingimprovingtemporalreasoning, mavromatis2021tempoqrtemporalquestionreasoning}.
These problems are often lengthy and involve complex temporal reasoning, yet still have clear correctness criteria that enable unambiguous assessment of critical gaps.
For instance, common failure patterns include BC/AD era confusion, systematic off-by-one errors, and inconsistencies in day-of-week arithmetic.
These issues are particularly evident and quantifiable using established benchmarks like Test of Time (ToT)~\cite{fatemi2025test}.
One example of a miscalculated time difference is provided in Appendix~\ref{sec:appndx_temporal}.
By investigating evaluation results to pinpoint \textit{domain-specific error patterns}, these patterns serve as seeds for synthesizing targeted data augmentation.
Through gap-augmented data synthesis and fine-tuning, we successfully boosted \textit{Granite-3.3-8B-Instruct} accuracy on ToT by over 20 percentage points (from 25.4\% to 48.9\%).


\paragraph{Fill-in-the-Middle (FIM).}
A concrete instance we encountered in practice that illustrates this category is Fill-in-the-Middle (FIM) code completion, a fundamental capability for code assistance tools, requiring models to predict missing code segments given surrounding prefix and suffix context.
We classify FIM as a well-defined task because its evaluation can be carried out in a fully deterministic manner using established benchmarks such as HumanEval-Infilling~\citep{fillInTheMiddle, chen2021codex}, where metrics like \emph{PASS@1} and \emph{PASS@10} offer precise and reproducible measurements.
The correctness of model outputs can be unambiguously determined since the generated code is directly executable, allowing objective verification.
Using \textit{Granite-3.3-8B-Instruct} as a test model, initial evaluations on HumanEval-Infilling revealed that the model struggled with this capability.
One such example is provided in Appendix~\ref{sec:appndx_fim}.
We conducted error pattern analysis and found that although generated code was largely correct, indentation and formatting were frequently misaligned.
This analysis helped identify the root cause of the failures, revealing a gap in the training data: the original dataset was split at logical code boundaries, so middle segments rarely contained indentation.
Consequently, the model never learned to reproduce proper formatting.
To address this, we regenerated the FIM dataset by randomly splitting code into prefix, middle, and suffix segments, thereby preserving natural indentation patterns.
This systematic approach yielded substantial improvements across all completion scenarios.
By combining targeted error analysis with dataset reconstruction, FIM performance improved significantly on a tested model.
Initial evaluations revealed \emph{PASS@1} scores as low as 0.15 for single-line tasks and even lower for multi-line (0.06) and random-span (0.015) cases.
After fine-tuning its base model (\textit{Granite-3.3-8B-Base}) on the regenerated dataset, single-line tasks improved by 0.57 points (from 0.15 to 0.72), multi-line tasks by 0.39 points (from 0.06 to 0.45), and random-span tasks by 0.545 points (from 0.015 to 0.56).
These gains confirm that structured error analysis and data regeneration can directly translate into measurable model improvements.


\subsection{Underspecified Tasks}
\label{sec:underspecified_use_cases}

In these settings, tasks may appear measurable but existing benchmarks or evaluation criteria are incomplete, misaligned, or fail to capture the true quality of outputs.
Debugging requires analyzing errors and refining the evaluation setup.


\paragraph{Natural Language to Mermaid Sequence (NL2Mermaid).}
In the case an ill-defined task where natural language is used to generate Mermaid~\citep{mermaid} sequence diagrams, debugging began with recognition that existing evaluations were insufficient to capture model performance.
Initial outputs revealed issues such as malformed syntax, missing activation handling, or ambiguous rendering of interactions.
A representative example is provided in Appendix~\ref{sec:appndx_nl2mermaid}.
Since no standardized benchmarks or evaluation criteria existed for this task, the first two phases of our framework (issue detection and evidence gathering) focused on clarifying what constitutes a ``good'' Mermaid diagram and constructing a reproducible evaluation.
This involved defining error categories, collecting seed examples, and expanding them with synthetic variants using \textit{SDG}~\cite{fms-dgt} to produce a dedicated benchmark, each pairing a natural language prompt with a gold-standard diagram.
The benchmark scores models across six criteria (i.e., syntax, mermaid-only output, logic, completeness, activation handling, error and status tracking) using two independent LLMaJ evaluators (\textit{DeepSeek-V3}~\citep{deepseekai2025deepseekv3technicalreport} and \textit{GPT-OSS-120B}~\citep{openai2025gptoss120b}).
This benchmark is formalized in MermaidSeqBench~\cite{shbita2025mermaidseqbench}.
By profiling model outputs across the new benchmark, we identified recurring weaknesses and systematically explored remedies.
One effective intervention was the use of in-context examples, which guided the LLM toward more consistent syntax and adherence to task conventions.
Results demonstrate that in-context examples substantially improved model performance: for instance, \textit{Llama-3.1-8B-Instruct}~\citep{grattafiori2024llama3herdmodels} improved from an overall score of 64.63 to 80.58, and \textit{Granite-3.3-8B-Instruct} from 46.96 to 73.90, validating both the benchmark and the debugging process.


\paragraph{Retrieval-Augmented Generation (RAG) QA.}
In some cases, poor performance arises from noisy benchmarks rather than the model itself.
Retrieval-Augmented Generation (RAG) Question-Answering (QA) benchmarks suffer from noise: reference answers (especially LLM-generated) may be unreliable, LLMaJ evaluations may diverge from human judgment, and evaluation criteria are unclear.
Noise detection begins with identifying performance gaps across tested models.
When several models failed in the same subset of a benchmark, the issue may originate from the benchmark rather than the models (see example in Appendix~\ref{sec:appndx_scribeflow}).
To investigate, model outputs can be compared against both subject-matter expert (SME) annotations and LLMaJ judgments.
Differences between SME and LLMaJ scores revealed judge noise, while comparisons across benchmarks showed that if a model performed well on similar datasets but poorly on one, the problem was likely due to that benchmark rather than the model.
Based on these findings, reference answers were re-examined, and the LLMaJ design was refined to better align with SME assessments.
Table~\ref{tab:rag_tall} summarizes model performance before and after noise resolution, using \textit{Llama-3.3-70B-Instruct}~\citep{grattafiori2024llama3herdmodels} as LLMaJ with evaluation aligned to SME annotations, unreliable metrics removed, and additional judges added.


\begin{table*}[t]
\centering
\small
\caption{Performance comparison of RAG QA tasks before and after resolving noise. Metrics:
  Correctness (answer accuracy), Completeness (information coverage), Adherence (instruction
  following), RL-F (retrieval-level F1 score), RB-LLM (retrieval-based LLM evaluation), and
  Conciseness (response brevity).}
\label{tab:rag_tall}
\resizebox{.9\linewidth}{!}{%
\begin{tabular}{llcccccc}
\toprule
\textbf{Model} & \textbf{Phase} & \textbf{Correctness} & \textbf{Completeness} & \textbf{Adherence} & \textbf{RL-F} & \textbf{RB-LLM} & \textbf{Conciseness} \\
\midrule
\multirow{2}{*}{Granite-3.2-8B-Instruct} &
  Before & 0.74 & 0.76 & 0.81 & - & - & - \\
& After  & 0.90 & 0.91 & -    & 0.95 & 0.85 & 0.23 \\
\midrule
\multirow{2}{*}{Llama-3.3-70B-Instruct} &
  Before & 0.71 & 0.72 & 0.91 & - & - & - \\
& After  & 0.87 & 0.89 & -    & 0.96 & 0.85 & 0.31 \\
\midrule
\multirow{2}{*}{Llama-3.1-8B-Instruct} &
  Before & 0.69 & 0.71 & 0.75 & - & - & - \\
& After  & 0.83 & 0.86 & -    & 0.93 & 0.79 & 0.20 \\
\midrule
\multirow{2}{*}{Llama-3.1-70B-Instruct} &
  Before & 0.70 & 0.70 & 0.84 & - & - & - \\
& After  & 0.84 & 0.85 & -    & 0.97 & 0.84 & 0.40 \\
\bottomrule
\end{tabular}
}
\end{table*}


\subsection{Agentic and Workflow-Level Tasks}
\label{sec:agentic_use_cases}

For multi-step or agent-based applications, issues often surface at the trajectory level: tool calls may be misaligned with user goals, reasoning may fail midway, or outputs may be inconsistent across runs.
Debugging here emphasizes trajectory analysis and prompt refinement, with iterative loops across phases  required to stabilize performance.

\paragraph{Site Reliability Engineering (SRE) workflow.}
To illustrate an agentic debugging scenario, we evaluated \textit{Granite-3.3-8B-Instruct} on an internal SRE workflow where an agent was tasked with diagnosing and remediating incidents in a Kubernetes environment and tested with models like \textit{GPT-OSS-120B} and \textit{Mistral Medium}~\citep{MistralAI2025Medium3}.
The agent received alerts, queried logs and metrics, and recommended mitigation steps such as scaling resources or restarting services.
Performance was evaluated using a rubric-based scoring system, where score of 6 indicates baseline performance (no improvement), 7 indicates minor improvement (e.g., clearer root cause attribution), while 8 indicates significant improvement (e.g., accurate propagation chains and effective remediation), and 9 indicates complete success (e.g., full resolution of the incident with minimal false positives).
The agent was guided by a structured system prompt that defined its role, goals, constraints, and available tools (e.g., \texttt{NL2Kubectl}, \texttt{NL2Metrics}), emphasizing structured reasoning and adherence to tool constraints.
A representative example is provided in Appendix~\ref{sec:appndx_agentic}.

The training data consisted of SRE incident logs and synthetic trajectories generated using \textit{SDG}, covering scenarios such as high-latency alerts, pod failures, network policy violations, and resource exhaustion.
First, \textit{hallucinated tool calls} emerged when the agent generated non-existent pod names or API invocations; these were mitigated by refining prompt structure with tool descriptions and constraints.
Second, \textit{trajectory pathologies} included redundant log queries, misaligned tool use, or premature escalation; trajectory profiling and adaptive thresholds helped flag such failures, while prompt refinements clarified escalation and retry strategies.
Third, \textit{dynamic thresholding} adapted error tolerances to reliability or relaxing constraints.

Evaluation focused on root cause identification accuracy, trajectory efficiency, hallucination rate, and output stability.
Post-optimization, the agent demonstrated improved generalization to unseen incident types (e.g., database connection failures) with minimal performance degradation on general tasks.
Table~\ref{tab:sre_debugging_results} summarizes improvements: clearer root cause attribution, more accurate propagation tracing, and fewer incorrect tool calls.

\begin{table}[h]
    \centering
    \small
    \caption{Performance comparison of SRE debugging tasks before and after refinement.}
    \label{tab:sre_debugging_results}
    \resizebox{.9\linewidth}{!}{%
    \begin{tabular}{p{0.8cm} p{3.4cm} p{1.9cm}}
        \toprule
        \textbf{Incident} & \textbf{Key Changes in Output}                          & \textbf{Score Trend} \\
        \midrule
        23                & Improved root cause attribution; aligned propagation chains & $\uparrow$ from $\sim$6 to 8 \\
        1                 & Trimmed root causes to cleaner causality                    & $\leftrightarrow$ held at 6 \\
        3                 & Additional propagation found in revised output              & $\uparrow$ from 7 to 8 \\
        \bottomrule
    \end{tabular}
    }
\end{table}

\section{Conclusion}

This paper introduced a four-phase framework for systematically debugging LLMs, spanning issue detection, evidence gathering, behavioral analysis, and iterative refinement.
We demonstrated the framework applies across three categories of tasks: well-defined (NL2GQL, Temporal Reasoning, FIM), underspecified (NL2Mermaid, RAG QA), and workflows (SRE).
These use cases show the framework supports debugging in diverse settings and provides a reproducible foundation for building LLM debugging infrastructure.
By integrating evaluation, diagnostics, and refinement into a closed loop, the framework moves beyond best practices for real-world LLMs.
This work establishes LLM debugging as an engineering discipline for transparency and reliability, offering a blueprint for diagnosis and improvement of LLM systems.
\section*{Limitations and Ethical Considerations}
\label{sec:limitations}

Our proposed framework and approach have several important limitations and ethical considerations.
First, while we demonstrate the methodology through use cases spanning well-defined tasks, underspecified tasks, and agentic workflows, these case studies are illustrative rather than exhaustive.
They cannot capture the full diversity of settings in which LLM debugging is required, and large scale empirical validation across domains, languages, and deployment environments remains an open direction.
Second, the decision to treat models as externally observable systems prioritizes generality and accessibility but limits diagnostic depth.
Without access to training data provenance, architectural details, or gradient-level signals beyond externally observable behavior, our approach cannot fully disentangle whether failures arise from model design, training distribution, or deployment conditions.
This restricts its explanatory power compared to complete white box analyses.
Finally, the framework makes use of LLMaJ and heuristic-based scorers for evaluation in underspecified and ill defined settings.
While these methods enable progress in the absence of standardized benchmarks, they also introduce potential biases, inconsistencies, and reproducibility challenges.
These concerns are especially salient for safety critical applications, where biased or unreliable evaluations may lead to harmful outcomes and where more rigorous, validated protocols are required.

\bibliography{llm-dot-debug}

\appendix


\section{Use Cases Details and Examples}
\label{sec:appndx_examples}


\subsection{Natural Language to GraphQL (NL2GQL) Use Case Details}
\label{sec:appndx_nl2gql}

Listing~\ref{lst:gql_prompt} provides one full prompt example for an NL2GraphQL task.
Listing~\ref{lst:gql_error} illustrates a faulty output we observed during evaluation.

\begin{lstlisting}[
    label=lst:gql_prompt,
    basicstyle=\ttfamily\scriptsize,
    frame=single,
    caption={Example of full prompt for the request: ``Give me the area km square and the hosts of the farm competition for city with id 1''.},
    captionpos=b,
    numbers=none,
    breaklines=true,
    breakindent=0pt,
    showstringspaces=false,
    linewidth=\columnwidth
]
You are an agent that is an expert in generating GraphQL queries. Your task is to generate the GraphQL query API request for a given the GraphQL reference schema and user instructions in natural language. You must structure your response with the following request format below, with the name of the function being called `query MyQuery { ... }` as the query name. You should not use any other name, other than `query MyQuery { ... }`. YOU MUST FOLLOW THIS FORMAT. 
SCHEMA:
```graphql
interface census_details{
  census_ranking: String
  population: Float
}
type City implements census_details{
  area_km_2: Float
  city_id: Int!
}
extend type City {
  farm_competition: [Farm_competition]
}
type Competition_record {
  competition_id: Int!
  farm: Farm
  farm_competition: Farm_competition
  farm_id: Int!
}
type Farm {
  farm_id: Int!
}
extend type Farm {
  competition_record: [Competition_record]
}
type Farm_competition {
  city: City
  competition_id: Int!
  competition_record: [Competition_record]
  host_city_id: Int
  hosts: String
}
type Query {
  city(city_id: Int!): City
}
```
Write a single graphql query which MUST start with `query MyQuery {`, with no explanation or additional text, just the query, to answer the following question:
Give me the area km square and the hosts of the farm competition for city with id 1.
\end{lstlisting}

Listing~\ref{lst:gql_GT} shows the correct expected output, which was achieved after fine-tuning the same model.
We use Low Rank Adaptation (LoRA) to fine-tune \textit{Granite-3.1-8B-Instruct}~\citep{granite2024granite} base model, using a train batch size of 4, 2 gradient accumulation steps, 2 epochs, a learning rate of $10^{-4}$, cosine scheduler type, and a max sequence length of 4,096.
The test and training data for this task were obtained from~\cite{kesarwani-etal-2024-graphql}.
The dataset includes 10,940 training triples spanning 185 cross-source data stores and 957 test triples over 14 data stores.
Each triple consists of a GraphQL schema, a GraphQL query operation, and a corresponding natural language query.
The dataset has been predominantly manually created, with natural language paraphrasing, and carefully validated, requiring approximately 1,200 person-hours.

\begin{lstlisting}[
    label=lst:gql_error,
    basicstyle=\ttfamily\scriptsize,
    frame=single,
    caption={Example of incorrect query generated by an open sourced LLM for the request: ``Give me the area km square and the hosts of the farm competition for city with id 1''.},
    captionpos=b,
    numbers=none,
    breaklines=true,
    breakindent=0pt,
    showstringspaces=false
]
{
  city(city_id: 1) {
    farm_competition(competition_id: 1) {
      host_city_id
      hosts
    }
  }
}
\end{lstlisting}

\begin{lstlisting}[
    label=lst:gql_GT,
    basicstyle=\ttfamily\scriptsize,
    frame=single,
    caption={Example of expected GraphQL query for the request: ``Give me the area km square and the hosts of the farm competition for city with id 1''.},
    captionpos=b,
    numbers=none,
    breaklines=true,
    breakindent=0pt,
    showstringspaces=false
]
query MyQuery {
  city(city_id: 1) {
    area_km_2
    farm_competition {
      hosts
    }
  }
}
\end{lstlisting}

As shown in Table~\ref{tab:nl2gql_debugging_results}, the initial test is a zero-shot test of the selected base model (\textit{Granite-3.1-8B-Instruct}) on the task.
The accuracy was around 20\%.
Performing empirical prompt optimization raised the accuracy to 27\%.
We executed a \textit{glass-ceiling test} by adding test data in the fine-tuning step.
The model reached 96\% accuracy.
Finally, \textit{vanilla fine-tuning} with the 10K data samples achieved an overall $\sim$48\% accuracy. 

\begin{table}[ht]
    \centering
    \small
    \caption{Performance results for \textit{Granite-3.1-8B-Instruct}~\citep{granite2024granite}, fine tuned using 10K data points~\citep{kesarwani-etal-2024-graphql}.}
    \label{tab:nl2gql_debugging_results}
    \resizebox{.5\linewidth}{!}{%
    \begin{tabular}{lc}
        \toprule
        \textbf{Test Variant} & \textbf{Accuracy} \\
        \midrule
        Zero Hop    & 82.71\%\\
        One Hop     & 53.81\% \\
        Two Hop     & 25.00\% \\
        FL Zero Hop & 41.94\% \\
        FL One Hop  & 35.53\% \\
        FL Two Hop  & 11.11\% \\
        Overall     & 47.77\% \\
        \bottomrule
    \end{tabular}
    }
\end{table}

As a proof of generalization check, we also verified that the fine-tuning has no negative effect on the standard base models' capabilities, i.e., it does not generate catastrophic forgetting on other generic tasks. We tested the original model (\textit{Granite-3.1-8B-Instruct}) and the LoRA fine-tuned version on the most prominent leaderboard benchmarks for LLMs, obtaining comparable results and observed no degradation in the  scores.


\subsection{Temporal Reasoning Use Case Details}
\label{sec:appndx_temporal}

While base models are capable of solving simple temporal problems, they frequently fail to reach the correct answer in specific categories.
Listings~\ref{lst:temporal-reasoning-user-prompt} and \ref{lst:temporal-reasoning-output} illustrate a representative user prompt and its erroneous response, respectively.
In this example, the base model was asked to compute the exact time difference between two timestamps.
The generated output incorrectly expressed the duration using a fractional day and mismatched hours, highlighting a recurring weakness in temporal reasoning tasks.

\begin{lstlisting}[
    label=lst:temporal-reasoning-user-prompt,
    basicstyle=\ttfamily\scriptsize,
    frame=single,
    caption={Prompt text for a temporal reasoning task.},
    captionpos=b,
    numbers=none,
    breaklines=true,
    breakindent=0pt,
    showstringspaces=false
]
Calculate the time difference between 5:00AM Nov 19, 2024 and 6:00AM Nov 25, 2024. List the answer using days, hours, and minutes.
\end{lstlisting}

\begin{lstlisting}[
    label=lst:temporal-reasoning-output,
    basicstyle=\ttfamily\scriptsize,
    frame=single,
    caption={Erroneous model output for the temporal reasoning task.},
    captionpos=b,
    numbers=none,
    breaklines=true,
    breakindent=0pt,
    showstringspaces=false
]
The time difference between 5:00 AM Nov 19, 2024, and 6:00 AM Nov 25, 2024, is 0.75 days, 6 hours, and 0 minutes.
\end{lstlisting}

To address this task, we materialized an ``Expert Assistant'' system prompt that mandates a structured response format.
Models are required to generate a detailed intermediate steps trace enclosed in \texttt{<think>} tags, followed by the final result in \texttt{<answer>} tags.
Listing~\ref{lst:temporal-reasoning-prompt} illustrates the full prompt structure, whereas Listing~\ref{lst:temporal-reasoning-GT} shows the desired correct behavior achieved after training on our curated synthetic data.

\begin{lstlisting}[
    label=lst:temporal-reasoning-prompt,
    basicstyle=\ttfamily\scriptsize,
    frame=single,
    caption={Example of a full prompt for a temporal reasoning task using an Expert Assistant persona.},
    captionpos=b,
    numbers=none,
    breaklines=true,
    breakindent=0pt,
    showstringspaces=false,
    linewidth=\columnwidth
]
You are an expert assistant specialized in temporal reasoning. Your task is to analyze events, times, durations, and their relationships to answer questions accurately. Always reason step by step, explicitly tracking time points, intervals, and order of events.

Answer Format:
<think>
{{Your step-by-step reasoning here}}
</think>
<answer>{{Final answer here}}</answer>
\end{lstlisting}

\begin{lstlisting}[
    label=lst:temporal-reasoning-GT,
    basicstyle=\ttfamily\scriptsize,
    frame=single,
    caption={Correct (desired) assistant output for the temporal reasoning task in Listing~\ref{lst:temporal-reasoning-user-prompt} showing the CoT reasoning chain.},
    captionpos=b,
    numbers=none,
    breaklines=true,
    breakindent=0pt,
    showstringspaces=false
]
ASSISTANT: 
<think>
1. From Nov 19, 5:00 AM to Nov 25, 5:00 AM is exactly 6 days.
2. From Nov 25, 5:00 AM to Nov 25, 6:00 AM is 1 hour.
3. Total duration is 6 days and 1 hour.
</think>
<answer>6 days, 1 hour, and 0 minutes</answer>
\end{lstlisting}

Listings~\ref{lst:temporal-reasoning-up2} and~\ref{lst:temporal-reasoning-sample2} show another example of a user prompt and the correct response, which we achieved after training on our curated synthetic data, respectively.

\begin{lstlisting}[
    label=lst:temporal-reasoning-up2,
    basicstyle=\ttfamily\scriptsize,
    frame=single,
    caption={Prompt text for a temporal reasoning task.},
    captionpos=b,
    numbers=none,
    breaklines=true,
    breakindent=0pt,
    showstringspaces=false
]
If today is 2024-02-28, the museum exhibition opened on 2022-11-15. How many months have passed between the opening date and today?
\end{lstlisting}

\begin{lstlisting}[
    label=lst:temporal-reasoning-sample2,
    basicstyle=\ttfamily\scriptsize,
    frame=single,
    caption={Correct expected assistant output for a calendar-based reasoning task in Listing~\ref{lst:temporal-reasoning-up2} addressing identified gaps.},
    captionpos=b,
    numbers=none,
    breaklines=true,
    breakindent=0pt,
    showstringspaces=false
]
ASSISTANT: 
<think>
First calculate the number of whole calendar months between the two dates. From 2022-11-15 to 2024-02-15 is exactly 15 full months (2 years = 24 months, then subtract the 9-month offset from November to February: 24 - 9 = 15). Since the end date, 2024-02-28, is after the 15-month anniversary (the 15th day of February), those 15 months are fully completed. Therefore, 15 months have passed between the opening date and today.
</think>
<answer>15</answer>
\end{lstlisting}

We use \textit{Supervised Fine Tuning (SFT)} to train the \textit{Granite-3.3-8B-Base}~\cite{granite2024granite} model using a learning rate of $10^{-4}$, cosine scheduler, and a max sequence length of 4,096.
The final dataset of 15,910 samples was synthesized using SDG~\cite{foundation-model-stack2025fmsdgt}.
Generation was seeded using concepts targeting the following identified failure trends:
(i) \textit{ToT-Arithmetic}~\cite{fatemi2025test} categories based on poor-performing sections of the ToT benchmark: Schedule (27\%), Trick (29\%), Duration (12\%), and Multi-op (8\%), and 
(ii) \textit{Observed Error Patterns} with specific gaps including ``BC/AD Transition'', ``Off-by-One/Two'', ``Backwards Calculation'', ``Day-of-Week'', ``Large Timespan Arithmetic'', ``Time Scaling'', and ``Overlap Logic''.

The dataset was curated through a three-stage filtering process of 72,000 raw samples from \textit{Mistral-Large} and \textit{GPT-OSS-120B} which included:
(i) consensus check between teacher models ensured accuracy,
(ii) target model answerability via retaining samples where the base model (e.g., in this case \textit{Granite-3.3-8B-Instruct}) failed to provide a correct answer, and
(iii) CoT quality using an LLMaJ (\textit{Llama-3.3-70B}) evaluation of reasoning coherence and complexity (score $\ge 7/10$).

Looking at the ablation results in Table~\ref{tab:temporal_detailed_results}, the biggest takeaway is that being picky with our data pays off.
By layering our filters—specifically checking if teachers agree (Filter 1), targeting things the model already struggles with (Filter 2), and ensuring the reasoning is high quality (Filter 3), we saw a steady climb in performance from 0.38 to 0.60 in the combined score
Interestingly, different filters help in different ways: targeting ``answerability'' (A1) provided a massive boost in the \textbf{Duration} category, while adding quality checks (A3) proved to be the secret sauce for more nuanced \textbf{Trick} questions.
Ultimately, our ``Full'' pipeline (A5) proves that when we combine all three filters, we get a model that is significantly more well-rounded and capable across the board than the baseline.

\begin{table*}[t]
\centering
\small
\caption{Temporal reasoning performance breakdown on different categories of ToT-Arithmetic~\cite{fatemi2025test} tasks. Ablations test different combinations of filters applied on the synthesized data: Filter 1 (Final Answer Match), Filter 2 (Granite Answerability), and Filter 3 (Judge CoT+Answer Quality).}
\resizebox{\linewidth}{!}{%
\begin{tabular}{l c c c c c c c c}
\toprule
\textbf{Model / Tasks} & \textbf{Combined} & \textbf{Add/Sub} & \textbf{Compare} & \textbf{Duration} & \textbf{Multi\_op} & \textbf{Schedule} & \textbf{Timezone} & \textbf{Trick} \\
& \tiny(n=1850) & \tiny(n=350) & \tiny(n=350) & \tiny(n=200) & \tiny(n=350) & \tiny(n=250) & \tiny(n=100) & \tiny(n=250) \\
\midrule
Llama-3.1-8B~\cite{grattafiori2024llama3herdmodels} & 0.36 & 0.47 & 0.81 & 0.11 & 0.06 & 0.15 & 0.54 & 0.36 \\
Granite-3.3-8B-Instruct~\cite{granite2024granite} & 0.38 & 0.48 & 0.84 & 0.14 & 0.14 & 0.17 & 0.48 & 0.33 \\
\midrule
\textit{Ablations (Granite-3.3-8B-Base-FT)} \\
A1: F2 (Answerability) & 0.51 & 0.67 & 0.63 & 0.51 & 0.44 & 0.05 & 0.50 & 0.67 \\
A2: F1+F2 (Match+Ans) & 0.58 & 0.63 & \textbf{0.94} & 0.47 & 0.51 & 0.24 & \textbf{0.63} & 0.47 \\
A3: F1+F3 (Match+Qual) & 0.54 & 0.60 & 0.67 & 0.41 & \textbf{0.58} & 0.10 & 0.61 & \textbf{0.78} \\
A4: F2+F3 (Ans+Qual) & 0.58 & \textbf{0.67} & 0.90 & 0.43 & 0.53 & 0.06 & 0.54 & 0.70 \\
A5: F1+F2+F3 (Full) & \textbf{0.60} & 0.65 & 0.93 & \textbf{0.48} & 0.49 & \textbf{0.27} & 0.57 & 0.70 \\
\bottomrule
\end{tabular}
}
\label{tab:temporal_detailed_results}
\end{table*}

We verified that fine-tuning improved task-specific capabilities without degrading base model performance across leaderboards.
Models maintained comparable results across leaderboards (Safety, RAG, Multi-lingual) while specifically improving accuracy in the \texttt{ToT\_Arithmetic} domain.


\subsection{Fill-in-the-Middle (FIM) Use Case Details}
\label{sec:appndx_fim}

In FIM tasks, the model must generate only the missing line of code at the marked location, without rewriting or altering the surrounding context.
Listings~\ref{lst:fim-prompt}, \ref{lst:fim-close-output}, and~\ref{lst:fim-close-desired} illustrate an example where the model is required to generate only the missing line of code (a function definition).

\begin{lstlisting}[
    label=lst:fim-prompt,
    basicstyle=\ttfamily\scriptsize,
    frame=single,
    caption={Prompt text for a Fill-in-the-Middle (FIM) task.},
    captionpos=b,
    numbers=none,
    breaklines=true,
    breakindent=0pt,
    showstringspaces=false
]
<fim_prefix>
from typing import List

def has_close_elements(numbers: List[float], threshold: float) -> bool:
    """ Check if in given list of numbers, are any two numbers
    closer than the given threshold. """
    <fim_suffix>
    for idx2, elem2 in enumerate(numbers):
        if idx != idx2:
            distance = abs(elem - elem2)
            if distance < threshold:
                return True
    return False
    <fim_middle>
\end{lstlisting}

\begin{lstlisting}[
    label=lst:fim-close-output,
    basicstyle=\ttfamily\scriptsize,
    frame=single,
    caption={Erroneous model output filling the missing line incorrectly.},
    captionpos=b,
    numbers=none,
    breaklines=true,
    breakindent=0pt,
    showstringspaces=false
]
for elem in enumerate(numbers):\n
\end{lstlisting}

\begin{lstlisting}[
    label=lst:fim-close-desired,
    basicstyle=\ttfamily\scriptsize,
    frame=single,
    caption={Correct output filling the missing line in the FIM task.},
    captionpos=b,
    numbers=none,
    breaklines=true,
    breakindent=0pt,
    showstringspaces=false
]
for idx, elem in enumerate(numbers):\n
\end{lstlisting}

Similarly to Listing~\ref{lst:fim-prompt}, we adopt the structured prompt format shown in Listing~\ref{lst:fim-structured-prompt}, where \texttt{<fim\_prefix>} and \texttt{<fim\_suffix>} delimit the prefix and suffix context, while \texttt{<fim\_middle>} marks the missing span to be predicted.
The model must emit only the missing middle segment such that:
(i) the code is syntactically valid and executable where applicable,
(ii) indentation/formatting is preserved (critical for Python and whitespace-sensitive languages),
and (iii) the completion is logically consistent with both prefix and suffix.

\begin{lstlisting}[
    label=lst:fim-structured-prompt,
    basicstyle=\ttfamily\small,
    frame=single,
    caption={FIM prompt template.},
    captionpos=b,
    numbers=none,
    breaklines=true,
    breakindent=0pt,
    showstringspaces=false
]
<fim_prefix>{prefix}<fim_suffix>{suffix}<fim_middle>{middle}
\end{lstlisting}

Fine-tuning was performed on \textit{Granite-3.3-8B-Base} with a maximum sequence length of 4,096 tokens, a learning rate of $10^{-5}$, and trained for 1 epoch.
We fine-tune on a regenerated multilingual FIM corpus designed to preserve natural indentation and realistic code boundaries:
(i) Python (150K samples), and
(ii) Other languages (200K total) covering: Java, C++, Go, JavaScript, Rust, and TypeScript.
Samples are randomly split into \texttt{prefix}, \texttt{middle}, and \texttt{suffix} segments.
The dataset includes \emph{single-line}, \emph{multi-line}, and \emph{random-span} cases.
We evaluate on HumanEval-Infilling~\citep{fillInTheMiddle, chen2021codex} using deterministic code execution with metrics \emph{PASS@1} and \emph{PASS@10}.

We assessed robustness by evaluating across varying span lengths, including single-line, multi-line, and random-span cases, to ensure the model handles different infilling complexities.
Additionally, we verified formatting fidelity (e.g., indentation and code style) and confirmed that the model correctly localized completions using the \texttt{<fim\_middle>} marker.
Error analysis was performed to identify common failure patterns such as misaligned indentation and logical inconsistencies.


\subsection{Natural Language to Mermaid Sequence Diagram (NL2Mermaid) Use Case Example}
\label{sec:appndx_nl2mermaid}

Listing~\ref{lst:uml-sequence-nl} illustrates one representative NL2Mermaid test case showing the natural language input specification for that task.
The listing shows the full input, including the purpose of the diagram, the main components, and the interactions in the form of an ordered sequence.

\begin{lstlisting}[
    label=lst:uml-sequence-nl,
    basicstyle=\ttfamily\scriptsize,
    frame=single,
    caption={Natural language specification for ``Uploading Documents with Secure Storage'' task.},
    captionpos=b,
    numbers=none,
    breaklines=true,
    breakindent=0pt,
    showstringspaces=false
]
Purpose: Uploading Documents with Secure Storage
Main Components: User, Mobile App, BFF, Azure AD
Interactions:
1. User Action: User uploads a document through the mobile app.
2. Mobile App: Sends the document along with the session token to the BFF.
3. The BFF validates the token with Azure AD.
4. On successful upload, the BFF returns a confirmation to the app. If the user is unauthorized or the file exceeds size limits, an appropriate error is returned.
\end{lstlisting}

Listing~\ref{lst:uml-sequence-mermaid} shows the corresponding erroneous Mermaid syntax generated by the LLM.
Here, the output diagram exhibited several issues that were recurrent across the dataset: inconsistent activation and deactivation markers, malformed or misplaced error-handling logic, and incomplete responses.

\begin{lstlisting}[
    label=lst:uml-sequence-mermaid,
    basicstyle=\ttfamily\scriptsize,
    frame=single,
    caption={Mermaid syntax illustrating a faulty ``Uploading Documents with Secure Storage'' flow, showing activation inconsistencies, incorrect error-handling, and incomplete actor responses.},
    captionpos=b,
    numbers=none,
    breaklines=true,
    breakindent=0pt,
    showstringspaces=false
]
sequenceDiagram
title Uploading Documents with Secure Storage

participant User
participant App as Mobile App
participant BFF
participant AAD as Azure AD

User->>App: Select document & upload
activate App
App->>BFF: POST /upload (file, token)
activate BFF
BFF->>AAD: validate (token)
activate AAD
AAD-->>BFF: 200 OK
deactivate App

alt success
  BFF-->>Application: 201 Created
  App-->>User: Done
  deactivate BFF
else unauthorized
  AAD-->>BFF: 401 Unauthorized
  BFF-->>App: 401
else too large
  BFF-->>App: 413 Payload Too Large
end
\end{lstlisting}


\subsection{RAG Benchmark Use Case Example}
\label{sec:appndx_scribeflow}

In Retrieval-Augmented Generation (RAG) tasks, the model should answer the question using only the provided information.
This involves:
(i) selecting relevant content,
(ii) handling multiple sources when more than one document is provided, and
(iii) managing uncertainty when the model cannot find the answer within the given information. 

Listing~\ref{lst:rag1} shows an example question with its reference answer. 
Listing~\ref{lst:rag2} shows a case where the human annotation aligns with scores given by an LLMaJ, while Listing~\ref{lst:rag3} shows a case where the human annotation does not fully align with the LLMaJ scores. 
In the latter example, the LLMaJ assigns a perfect score even though some information is missing.

\begin{lstlisting}[
    label=lst:rag1,
    basicstyle=\ttfamily\scriptsize,
    frame=single,
    caption={Example question and reference answer (documents truncated due to page limits).},
    captionpos=b,
    numbers=none,
    breaklines=true,
    breakindent=0pt,
    showstringspaces=false
]
Q: How are estimation methods applied in developing medical costs payable estimates?
A: In developing medical costs payable estimates, different estimation methods are applied depending on the month for which incurred claims are being estimated.
\end{lstlisting}

\begin{lstlisting}[
    label=lst:rag2,
    basicstyle=\ttfamily\scriptsize,
    frame=single,
    caption={Case where human annotation aligns with scores given by an LLMaJ.},
    captionpos=b,
    numbers=none,
    breaklines=true,
    breakindent=0pt,
    showstringspaces=false
]
Model Response 1: In developing medical costs payable estimates, the company applies different estimation methods depending on the month for which incurred claims are being estimated. For the most recent two months, they estimate claim costs incurred by applying trend factors, which represent an estimate of claims expense based on recent claims expense levels and healthcare trend levels. For prior months, the company uses completion factors, representing the average percentage of total incurred claims paid through a given date after being incurred, based on historical data. These completion factors are applied to claims paid through the period-end date to estimate the ultimate claim expense incurred for the period. Actuarial estimates of incurred but not paid claim liabilities are then determined by subtracting the actual paid claims from the estimate of the ultimate incurred claims.

Human annotation: Discusses both "most recent two months" and "prior months".
LLMaJ scores: correctness 1.0, completeness 1.0
\end{lstlisting}

\begin{lstlisting}[
    label=lst:rag3,
    basicstyle=\ttfamily\scriptsize,
    frame=single,
    caption={Case where human annotation does not fully align with scores given by an LLMaJ.},
    captionpos=b,
    numbers=none,
    breaklines=true,
    breakindent=0pt,
    showstringspaces=false
]
Model Response 2: In developing medical costs payable estimates, different estimation methods are applied depending on the month for which incurred claims are being estimated. For the most recent two months, the estimate of claim costs incurred is based on applying actuarial models that require management to make certain judgments. These include medical cost per member per month trend factors and completion factors, which consider assumptions such as the time from date of service to claim receipt, the impact of actual care activity, and processing cycles.

Human annotation: Missing completion factor for "prior months".
LLMaJ scores: correctness 1.0, completeness 1.0
\end{lstlisting}


\subsection{Agentic Tooling Use Case for SRE}
\label{sec:appndx_agentic}

Listing~\ref{lst:sre-prompt} presents a representative example of an SRE workflow evaluated under our systematic debugging framework.
The task involves diagnosing and remediating a Kubernetes incident, where the agent must identify root causes, trace fault propagation chains, and recommend mitigation steps.
The example below illustrates a typical prompt and the agent's reasoning trajectory, showing how structured refinement improves accuracy and reduces hallucinations.

\begin{lstlisting}[
label=lst:sre-prompt,
basicstyle=\ttfamily\scriptsize,
frame=single,
caption={System prompt for the SRE diagnosis agent, defining its role, goals, constraints, and available tools. The prompt emphasizes structured reasoning, tool constraints, and the need to avoid assumptions.},
captionpos=b,
numbers=none,
breaklines=true,
breakindent=0pt,
showstringspaces=false
]
You are sre_diagnosis_agent, a smart and capable tool-using agent. Your primary goal is to diagnose IT incidents in Kubernetes environments by identifying fault propagation chains and root causes.
Available tools:
GetAlerts Tool: Retrieves current alerts.
NL2Kubectl Tool: Executes kubectl commands (e.g., describe pods, get logs).
NL2Metrics Tool: Accesses Prometheus metrics.
NL2Traces Tool: Gathers Jaeger traces.
NL2Logs Tool: Retrieves Loki logs.
Follow this format:
Thought: [Your reasoning]
Action: [Tool name]
Action Input: {"query": "..."}
Observation: [Result]
Example procedure:
Get current alerts.
Investigate affected pods/services using NL2Kubectl.
Analyze metrics/traces/logs for root causes.
Iterate until all root causes are identified.
Diagnose the incident: "Alert: High latency in frontend service in app-xyz namespace."
\end{lstlisting}

Listing~\ref{lst:sre-trajectory} shows a representative trajectory of the agent's reasoning and tool use.
The agent begins by gathering alerts, then iteratively investigates pods, logs, and network configurations to isolate the root cause (in this case, a misconfigured service port).

This trajectory underscores the value of structured agent reasoning in diagnosing complex IT incidents.
By adhering to tool-use constraints and prompt structure—avoiding hallucinations, redundant actions, and unsupported assumptions, the agent systematically narrows potential root causes and produces reliable diagnoses.
The example also illustrates common challenges in SRE workflows, such as misconfigured services or restrictive network policies, and demonstrates how the debugging framework guides resolution through evidence gathering and refinement.

\begin{lstlisting}[
label=lst:sre-trajectory,
basicstyle=\ttfamily\scriptsize,
frame=single,
caption={Agent trajectory for diagnosing high latency in the frontend service, illustrating iterative refinement and tool use.},
captionpos=b,
numbers=none,
breaklines=true,
breakindent=0pt,
showstringspaces=false
]
Thought: High latency alert in frontend. I will check the pod status.
Action: NL2Kubectl Tool
Action Input: {"nl_query": "describe pod frontend-xyz-123 in app-xyz"}
Observation: Pod is Running, but logs show connection timeouts to api-service.
Thought: The issue may be with api-service. I will check its configuration.
Action: NL2Kubectl Tool
Action Input: {"nl_query": "describe service api-service-xyz in app-xyz"}
Observation: Service exposes port 8080, but the app expects port 9090.
Thought: The root cause is a port mismatch. I will update the service.
Action: NL2Kubectl Tool
Action Input: {"nl_query": "update service api-service-xyz to expose port 9090"}
Observation: Service updated. Latency alert resolved.
\end{lstlisting}

To address the issue, decoding parameters were set to ensure deterministic reasoning (temperature=0.2, top-$p$=0.9, max tokens=1,024), and fine-tuning, where applied, used LoRA (rank=8, $\alpha$=16) with a learning rate of $5\cdot10^{-5}$ to maintain efficiency and avoid overfitting.
The data mixture for fine-tuning included 80\% SRE-specific incident logs and 20\% generic IT operations documentation to balance task-specific and general knowledge. 
Finally, the agent achieved a markedly lower hallucination rate and reduced repeat action frequency, reflecting measurable gains in both efficiency and precision, as presented in Section~\ref{sec:agentic_use_cases}.

\end{document}